\documentclass{sigchi}

\usepackage{balance}       
\usepackage{graphics}      
\usepackage[T1]{fontenc}   
\usepackage{txfonts}
\usepackage{mathptmx}
\usepackage[pdflang={en-US},pdftex]{hyperref}
\usepackage{color}
\usepackage{booktabs}
\usepackage{textcomp}

\usepackage{microtype}        
\usepackage{ccicons}          

\usepackage{todonotes}

\def\plaintitle{Detecting gender differences in perception of emotion in crowdsourced data}

\def\emptyauthor{}
\def\plainkeywords{affective computing; perception; crowdsourcing; gender differences; data in the wild; hypothesis testing}

\makeatletter
\def\url@leostyle{%
  \@ifundefined{selectfont}{
    \def\UrlFont{\sf}
  }{
    \def\UrlFont{\small\bf\ttfamily}
  }}
\makeatother
\def\pprw{8.5in}
\def\pprh{11in}

\definecolor{linkColor}{RGB}{6,125,233}
\hypersetup{%
  pdftitle={\plaintitle},
  pdfauthor={\emptyauthor},
  pdfkeywords={\plainkeywords},
  pdfdisplaydoctitle=true, 
  bookmarksnumbered,
  pdfstartview={FitH},
  colorlinks,
  citecolor=black,
  filecolor=black,
  linkcolor=black,
  urlcolor=linkColor,
  breaklinks=true,
  hypertexnames=false
}

\begin{document}

\title{\plaintitle}

\numberofauthors{3}
\author{%
  \alignauthor{Shahan Ali Memon\\
    Hira Dhamyal\\
    \affaddr{Carnegie Mellon University}\\
    \affaddr{Pittsburgh, USA}\\
    \email{$\{$samemon,hyd$\}$@cs.cmu.edu}}\\
  \alignauthor{Oren Wright\\
  Daniel Justice\\
  Vijaykumar Palat\\
  William Boler\\
    \affaddr{Software Engineering Institute}\\
    \affaddr{Carnegie Mellon University}\\
    \affaddr{Pittsburgh, USA}\\
    \centerline{\email{$\{$owright,dljustice,vpalat,wmboler$\}$@sei.cmu.edu}}}
  \alignauthor{Bhiksha Raj\\
    Rita Singh \\
    \affaddr{Carnegie Mellon University}\\
    \affaddr{Pittsburgh, USA}\\
    \email{$\{$yandongw,bhiksha,rsingh$\}$@cs.cmu.edu}}
}

\maketitle

\begin{abstract}
Do men and women perceive emotions differently? Popular convictions place women as more emotionally perceptive than men. Empirical findings, however, remain inconclusive. Most prior studies focus on visual modalities. In addition, almost all of the studies are limited to experiments within controlled environments. Generalizability and scalability of these studies has not been sufficiently established. In this paper, we study the differences in perception of emotion between genders from speech data in the wild, annotated through crowdsourcing. While we limit ourselves to a single modality (i.e. speech), our framework is applicable to studies of emotion perception from all such loosely annotated data in general. Our paper addresses multiple serious challenges related to making statistically viable conclusions from crowdsourced data.
Overall, the contributions of this paper are two fold: a reliable novel framework for perceptual studies from crowdsourced data; and the demonstration of statistically significant differences in speech-based emotion perception between genders.
\end{abstract}

\ccsdesc[500]{Information systems~Crowdsourcing}
\ccsdesc[300]{General and reference~Empirical studies}
\ccsdesc[300]{Mathematics of computing~Resampling methods}

\keywords{\plainkeywords}

\printccsdesc

\section{Introduction}

One of the main goals of affective computing is to design computer systems that can act as emotionally \emph{sentient agents} \cite{mcduff2018designing} i.e. machines that are emotionally intelligent. A basic component of emotional intelligence is the ability to \emph{perceive}, and interpret emotions \cite{mayer2008human}. It is important to emphasize that perception, while correlated to emotional \emph{expression}, is different from it. This distinction can be best explained by the modified version of Brunswik's lens model proposed in \cite{scherer2003vocal}. There are three main stages in this model: the encoding, the transmission, and the decoding of emotions. Encoding is the process where individual conveys their internal state by modifying their communicative channel. Decoding is the process where another individual makes an inference about the state of the first individual. The cues that are encoded and the cues that are decoded may differ based on the noise in the transmission. When studying expression, the focus is on how the emotions were encoded, and the primary subject of study is the encoder. On the other hand, perception deals with how the emotions were interpreted or decoded, and, hence, the focus is on the decoder. To create emotionally intelligent machines that can interact with humans, understanding how humans perceive emotions is a crucial first step. Not only that, but to model a computational analogue of a human, it is imperative to understand how the perception of emotion \emph{varies}, if at all, across individuals. 

The topic of variation in emotion perception has been the source of long-standing debates among psychologists and evolutionary scientists for years. One of the commonly studied theories in this regard is the difference in perception of emotion between genders. The most robust gender stereotype in this regard is that ``women are more emotionally perceptive than men'' \cite{1,2,3,4,5,6,7,8}. The empirical findings, however, have been inconsistent with such popular convictions \cite{8}. 

Because the visual system is arguably considered as a primary mode of emotion perception, the study of facial expressions has been the fulcrum of most of the background literature \cite{hwang2015evidence}. The perception of emotion, however, extends far beyond the visual system. A second important modality in the process of emotion perception, and one that is the focus of this paper, is the auditory system. 
The background literature in the context of emotion perception in speech can be divided into two groups. One group of studies focuses on how the acoustic characteristics of voice affect the perception of emotion, examples of which include\cite{speechpercept11,speechpercept12,speechpercept13,speechpercept14,ahrens2014gender,schirmer2004gender}. In these studies, the characteristics of the voice (e.g. pitch, intonation, duration, and loudness), and not necessarily of the subjects, are the center of interest. The second group, in which this paper falls, is one wherein perceptions are studied across groups of individuals (e.g. male versus female, old versus young). Here the focus is not on the acoustics of the sound, but rather on the characteristics of the subjects. The study of perception of emotion in speech across genders has been studied multiple times \cite{speechpercept21,hall1978gender}. However, obtained results are mixed, bounded, or inconclusive \cite{lausen2018gender}. 

We identify two major shortcomings in the historical approaches. The first shortcoming is related to the quality and quantity of the stimuli. Most of these studies focus on a limited stock of voice samples, and use ``acted'' or ``elicited'' samples rather than natural. The second major shortcoming is that these studies are conducted on small scales in controlled environments. While this ensures that the quality of data is high, it limits the generalizability of the conclusions made from such experiments. Because the sample size of the subjects is typically small, this also makes the statistical power of the experiments relatively weak \cite{cohen2013statistical}.

In this work, we propose a scalable approach to the study of emotion perception applied to the modality of speech. We use speech data in the wild to collect stimuli with naturally occurring emotions. We take further steps to ensure the quality of the stimuli is maintained. Furthermore, we use crowdsourcing for collecting perceptual cues from thousands of subjects. While this approach scales relatively better in comparison to previous studies, it comes with some concomitant trade-offs. One of the biggest challenges of crowdsourcing is quality control as there is limited to no control over who responds, how they respond, and under what conditions they respond. In such a situation, a reasonable approach is to collect as much data as possible, without fully trusting the participants and the data. However, once the data are collected, accounting for anomalies in the data is an important but difficult problem, and has been articulated in many other contexts \cite{garcia2016challenges,gadiraju2017crowdsourcing}.

Because our data are composed of thousands of audio stimuli to achieve better generalizability, it is not possible to simulate a lab setting and make each rater rate all the audio samples. This creates a data record of annotations where raters have annotated only a small sample of partially overlapping data. If gender differences in perception are to be studied, data cannot be cleanly separated into male and female groups due to varying number of annotations by each rater and by each group. This creates rater-specific biases in the data, and poses great challenges for hypothesis testing as standard statistical tests cannot be directly applied. 

To overcome these challenges, We have adapted hypothesis testing for raters in such crowdsourced setting, to be able to deal with varying distributions of raters, where some occur more frequently than others; the only assumption being good-faith raters -- an assumption that we've tried to realize through design filters. We show how we can formulate our hypotheses based on appropriate filtering, re-sampling and conditioning of the data while deriving relevant statistical tests that can account for intra-rater variances and still have high power. We envision that our proposed framework will be extended in the future to study perception of emotion on scale.

\section{Related Work}
Perception of emotion in speech has been studied in many contexts: age, culture, gender, etc. Background research can be weakly categorized into two groups: 1) research focused on studying the changes in perception based on the changes in content of the vocal stimuli; and 2) research on studying the changes in perception between groups irrespective of the properties of the vocal stimuli. 

For speech emotion perception, most of the background literature falls under the former group. 
In this regard, \cite{schmidt2016perception} studies the relationship between the valence and arousal ratings on utterances and the acoustic parameters like pitch, intensity and articulation rate of the utterance causing the change in perception between different age groups. Many studies have focused on cultural differences in both the speakers and the raters \cite{scherer2001emotion,albas1976perception,bachorowski1999vocal}. \cite{sawamura2007common} performs study on people from three different backgrounds, Americans, Japanese and Chinese listening to utterances from Japanese children's saying the same thing in different emotions. One of the results of this study said that the identification rate of the emotion in the utterance is slightly higher for native-language listeners than for the non-native ones. This shows the cultural influence on perception. 

In terms of the studies on differences in speech emotion perception between genders, most of the studies \cite{speechpercept11,speechpercept12,speechpercept13,speechpercept14,ahrens2014gender,schirmer2004gender} focus on the acoustic properties and content of the vocal stimuli. \cite{schirmer2004gender} analyses the lateralization of word processing and emotional prosody processing between the two genders when provided with speech with emotional content. 

The most relevant studies to this paper include \cite{speechpercept21,lausen2018gender} where authors study the changes in perception of emotion in speech between gender groups based on different sets of stimuli. These studies, however, only focus on small sample of subjects, employ discrete emotions, or use acted stimuli.

In our work, we specifically focus on the differences in speech emotion perception between genders. We study these differences irrespective of the acoustic characteristics of the vocal stimuli, and only by comparing the evaluations of different gender groups. Unlike most previous research, we scale our perception test by including thousands of subjects and audio stimuli. We also build a novel and robust framework for such large scale studies for future researchers to follow and build upon.

\section{Data Description}
We divide the database collection process into two main parts: 1) utterance collection, and 2) annotation collection. The following sections describe each of the two parts in detail.

\subsection{Terminology}

For the purposes of this paper, we shall identify a single audio stimulus as an `utterance', the response to the stimulus as an `annotation', and the subject responding to the stimulus as the `rater'. 

\subsection{Utterance Collection}

A useful speech emotion database consists of short, clean, single-speaker clips with extemporaneous speech content that spans a wide range of emotions. Most available recordings in the literature are scripted, emotionally neutral, and feature music and background noise. For this study, we attempt to use data in the wild by selecting sources that contain extemporaneous speech and a wide range of emotions. Furthermore, we use appropriate pre-processing tools to remove degraded or noisy clips. 

\begin{figure*}[ht]
  \centering
  \scalebox{0.20}{
    \includegraphics{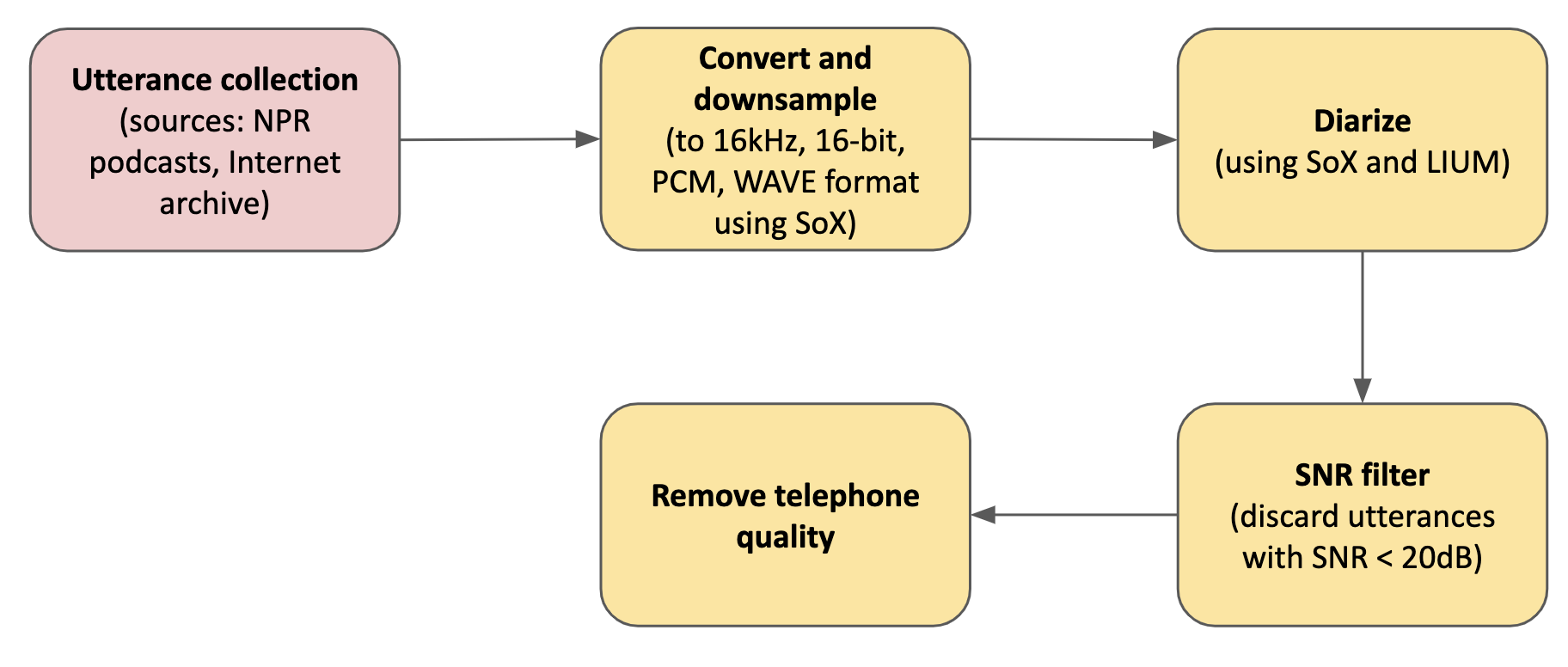}}
  \caption{Flow diagram for the utterance collection and the steps taken for pre-processing of utterances.}
  \label{fig:process}
\end{figure*}

\subsection{Source}

All the utterances are collected from NPR podcasts \cite{npr}, and television programs hosted by the Internet Archive \cite{internetarchive}. NPR podcasts include ``Serial", ``Radiolab", ``Wait Wait... Don't Tell Me!", and ``Car Talk". The Internet Archive data comes from the ``Understanding 9\/11" television news archive \cite{911} which contains 3,000 hours of TV news from 20 channels over 7 days. This news archive is specifically chosen with the hypothesis that the resultant utterances would include a wide range of emotional content.

\subsection{Pre-processing}
To ensure short, clean, single-speaker, and high quality utterances, we apply rigorous pre-processing controls before publishing the utterances to the raters. The audio utterances are first downsampled to 16 kHz, 16-bit, PCM, waveform audio file format files using SoX \cite{sox}. To ensure single-speaker utterances, they are further diarized using LIUM \cite{pypi}, and segmented into 3-10 second clips using SoX \cite{sox}. To remove utterances with high noise, we apply WADA-SNR filter \cite{kim2008robust} and discard clips with SNR of less than 20dB. As a final step, all the telephone quality audio files are also removed.

\subsection{Annotation Collection}

The pre-processed utterances were then uploaded to Amazon Mechanical Turk \cite{turk2012amazon} for raters to evaluate using three dimensional model of emotion described in terms of valence (or pleasure), arousal (or activation) and dominance. Based on the assumption that the participants of the study would be unaware of the affective dimensions, we employed AffectButton \cite{broekens2013affectbutton} as a design component of our study. To the best of our knowledge, this is the first study to use the AffectButton as a tool for labeling the perceptual cues. AffectButton allows a rater to project their perceptual cues onto a visual space using a visual emoticon. Projecting the perception of emotions onto a visual space is much more user-friendly and convenient for raters than to do it using a numerical scale \cite{broekens2013affectbutton}. The visual space of the AffectButton varies on a continuum by traversing it using a mouse. Each position on that space gives a unique valence, arousal and dominance value. The valence values change from -1 to 1 as the cursor moves from left to right, the arousal values change from -1 to 1 as the cursor moves away from the center of the space, and the dominance values change from -1 to 1 as the cursor moves from bottom to top. The use of AffectButton allows for a rater to conveniently mark their perceptual cues on a wider range without a need of an explanation for affective dimensions. Further details on the use of the three-dimensional emotion model, and the AffectButton can be found in the supporting information section in \nameref{S1_Appendix} and \nameref{S2_Appendix} respectively.

\begin{figure}[!ht]
  \centering
  \scalebox{0.22}{
    \includegraphics{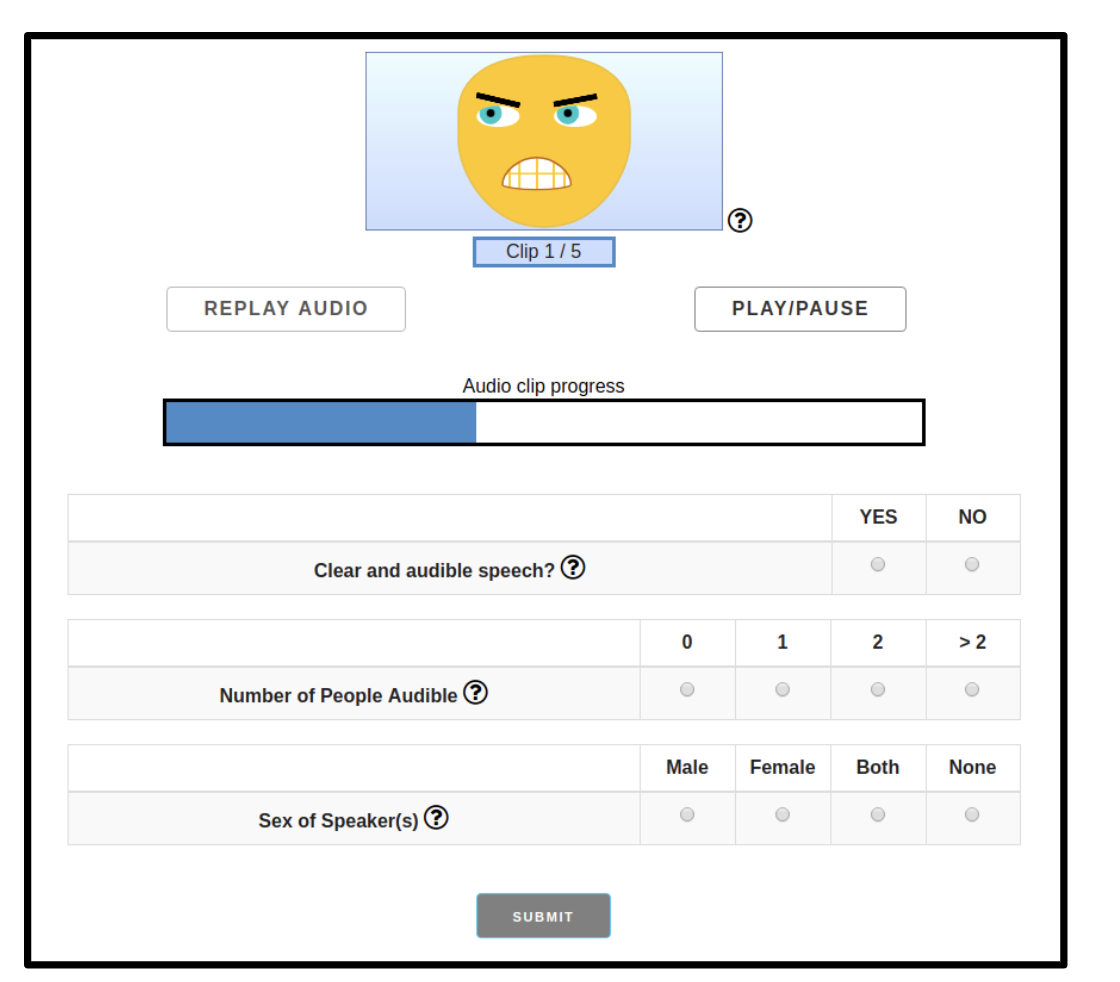}}
  \caption{Annotation tool used for emotion evaluation. The emotional content can be evaluated in terms of the three affective
dimensions: valence, arousal and dominance using the AffectButton.}
  \label{fig:app}
\end{figure}

Given the AffectButton, each rater listened to an audio utterance, and described the perceived emotion by dragging the mouse to an appropriate position on the AffectButton. Each rater was also asked to mark the sex of the speaker, if the utterance had multiple speakers, and if it was noisy. Each session comprised of 5 utterances. At the beginning of each session, the rater was asked demographic information such as sex, age, language, and country of origin. Figure \ref{fig:app} shows the main page for our deployed application.

\subsection{Annotation Filtering}

At the end of data collection, we performed the following steps for ensuring high quality annotations:
\begin{itemize}
    \item We removed all utterances with less than 5 annotations.
    \item We removed all utterances with only male or only female annotators.
    \item To remove bad faith actors, we removed all sessions where the variance in the annotation values of valence, arousal, or dominance was zero.
    \item We removed all the raters and their annotations who were inconsistent in their response to demographic information across sessions.
\end{itemize}

\subsection{Data statistics}
At the end of filtering, we were left with 163301 annotations comprising 17600 utterances, and 3343 raters. The 3343 raters covered 32 languages, 59 age values, and 80 countries. Out of 3343 raters, 1463 identified themselves as male, and 1880 as female. Average number of annotations per utterance was 9.28, whereas average number of annotations per rater was 48.85. Detailed statistics on this data are presented in Table~\ref{tab:datastatistics}.
We make this annotation data available to the public for future research prospects.


\begin{table}[ht]

\begin{tabular}{|c|c|} 
\hline
Statistic & Value  \\
\hline
\hline
Length of dataset & $\approx$ 28 hours\\ 
\hline
Average duration of an utterance & $\approx$ 5.76 seconds\\ 
\hline
\# of utterances & 17600\\
\hline
\# of annotations & 163301\\
\hline
\# of raters & 3343\\
\hline
\# of male raters & 1463\\
\hline
\# of female raters & 1880\\
\hline
\# of age values of raters & 59 \\
\hline
\# of languages covered by raters & 32\\
\hline
\# of countries covered by raters & 80\\
\hline
\# of annotations per rater on average & 48.85\\
\hline
\# of annotations per utterance on average & 9.28\\
\hline
Speaker gender distribution by majority vote: &  \\
Male & 13100\\
Female & 2251\\
Both & 1809\\
Neither & 239 \\
NA & 201\\
\hline
\hline
\end{tabular}
\caption{This table shows the statistics for the annotation data we make available for the public for future research projects. `NA' as a value means the data was not available.}
\label{tab:datastatistics}
\end{table}

\section{Challenges of Crowdsourced Data}
As mentioned earlier, collecting data through crowdsourcing brings numerous challenges with itself. 
When hypothesizing about the genders of the raters, many variables need to be controlled. One of the properties of the data is that an utterance is annotated by many raters. These raters are random for each utterance. Ideally, each rater would have annotated each utterance. However, the data at hand is very sparse. Since a random set of females and males have annotated an utterance and there exists an uneven distribution of the number of utterances each rater has annotated, the individual biases need to be taken into account. If this is not taken into account then instead of capturing the general trend of perception among the genders, we might capture the trend of dominant male raters versus dominant female raters.

As mentioned earlier, the data are very sparse. Sparsity of the data has been dealt with in many ways in literature. \cite{kang2013prevention} talks about the general ways to deal with missing data. Most common way of dealing with missing data is to delete those entries where the data are not complete. We however cannot do this because the data are incomplete for every utterance and so we would end up loosing all the utterances. 

\cite{guo2017comparative} talks about the missing imputation method using expectation maximization to estimate the missing values in data. \cite{guo2017comparative} also introduces paired data test statistics which take into account the unpaired data. However, the methods mentioned cannot be applied to our situation. We do not have one female rater and one male rater and some missing values in those. We have a set of female and male raters, where number of gender specific raters is different. Therefore we believe that the partially paired test statistics cannot be directly applied here. 

To tackle the above challenges, we reformulate our problem. We resample our data in a way which accounts for the rater biases, to reduce the missingness of the data. 
To create a viable sampling strategy, we created all possible (male,female) rater pairs from the data, each pair representing the list of annotations for each utterance which the pair co-annotated. We then randomly sampled each pair, and randomly picked an utterance without replacement. This process was repeated until all utterances were exhausted. This ensured that each utterance only had two annotations. This effectively removed the condition where the rater and utterance pair was repeated multiple times in the data. 

The sampling process yielded 35200 annotations comprising 17600 utterances, and 3167 raters. The 3167 raters covered 31 languages, 59 age values, and 76 countries. Out of 3167 raters, 1396 identified themselves as male, and 1771 as female. Number of annotations per utterance was 2, whereas average number of annotations per rater was 11.11.

\section{Hypothesis Testing}
We analyze two questions: (1) How do the two genders differ in perceiving emotion (2) How does a specific gender's perception differ based on the speaker's gender in a given utterance. 
In the first test, the data can be paired into subjects, namely utterances, which are annotated by both females and males. 
In the second, we have unpaired data where the ratings are done by, say, only females and the utterances contain both male and female speakers. 

In the following subsections we present how we formulate the two problems and the test statistics we derive. 

\subsection{Paired test}
The problem at hand can be formulated as follows: we are given a collection of utterances, each rated (for valence, activation and dominance) by one male and one female rater. We must determine if the two genders rate the utterance differently or not. The null hypothesis is that they have no difference in their ratings. 

Formally stated, let $x_{fi}$ and $x_{mi}$ be the ratings assigned to the $i^{\rm th}$ utterance by its female and male raters. Let $d_i = x_{fi} - x_{mi}$; i.e the difference between their ratings. 
The null and alternate hypotheses are:
\begin{align*}
H_0 : \mathbb{E}\left[d_i\right] = 0 \\
H_A : \mathbb{E}\left[d_i\right] \neq 0
\end{align*}
To evaluate the hypotheses, we will compute a $z$ statistic, based on the empirical average of $d_i$.
According to the Central Limit Theorem, $d_i$ follows a normal distribution. 
Let $\bar{d}$ be the empirical estimate of $\mathbb{E}\left[d\right]$. 
Under the assumption that $d_i$ is normally distributed, the test statistic follows a student-t distribution. Following the rules of the standard t-test,  we will compute the statistic $\frac{\bar{d}-\mathbb{E}\left[\bar{d}\right]}{\sqrt{\sigma_{\bar{d}}^2}}$, where $\sigma_{\bar{d}}^2$ is the variance of $\bar{d}$.  Under the null hypothesis $\mathbb{E}\left[\bar{d}\right]=0$. The variance of $\bar{d}$ must be carefully derived to account for the vagaries of our data.

Note that, in given situation, a rater may repeat multiple times across recordings. Different raters may repeat different numbers of times.  To account for the rater repetition, the variance of individual raters must be taken into account. Explicitly accounting for this fact, we derive $\sigma_{\bar{d}}^2$, and the corresponding t statistic as follows under $H_0$. First we define
\begin{equation*}
\bar{d}  = \frac{1}{N} \sum_i d_i 
\end{equation*}
Under $H_0$ $\mathbb{E}\left[\bar{d}\right]=0$. 
\begin{align}
    \sigma_{\bar{d}}^2 & = \mathbb{E}\left[\bar{d}^2\right] - (\mathbb{E}\left[\bar{d}\right])^2 \nonumber \\
    &= \mathbb{E}\left[\left(\frac{1}{N}\sum_id_i\right)^2\right] - \left(\mathbb{E}\left[\bar{d}\right]\right)^2 && \mathbb{E}\left[\bar{d}\right] = 0, \text{under } H_0\nonumber \\ 
    & = \mathbb{E}\left[\left(\frac{1}{N}\sum_id_i\right)^2\right] \nonumber \\
    & = \frac{1}{N^2} \left( \mathbb{E}\left[\left(\sum_id_i\right)^2\right] \right) \nonumber \\
    & = \frac{1}{N^2}\sum_i\mathbb{\mathbb{E}}\left[d_i^2\right] + \frac{1}{N^2} \sum_{i\neq j}\mathbb{E}\left[d_id_j\right] \nonumber \\ 
    \label{eq:stat1}
    & = \frac{1}{N}\sigma_{d}^2 + \frac{1}{N^2}\sum_{i\neq j}\mathbb{E}\left[d_id_j\right] 
\end{align}
where $\sigma_d^2$ is the variance of $d_i$. In the final step above we've utilized the fact that $\mathbb{E}\left[d\right]=0$, and hence $\mathbb{E}\left[d^2\right]=\sigma_d^2$

The second term to the right comprises the sum of $N(N-1)$ terms of the kind $\mathbb{E}\left[d_id_j\right]$. Each $(d_i$,$d_j)$ pair potentially involves {\em four} raters, the male and female rater for the $i^{\rm th}$ utterance, and the male and female raters for the $j^{\rm th}$ utterance. We can generally assume that separate raters are statistically independent of each other. 

When all four raters are different the ratings are independent of each other, and the term $\mathbb{E}\left[d_id_j\right]$ is zero. However, in our scenario either the male or the female rater (or both) may repeat across the utterance pair. In order to deal with this, we must explicitly compute $\mathbb{E}\left[d_i d_j\right]$ and consider all the various scenarios.

The term $\mathbb{E}\left[d_id_j\right]$ can be explicitly written down as the product of the differences of the ratings.
\begin{align*}
    \mathbb{E}\left[d_id_j\right]&= \mathbb{E}\left[(x_{mi} - x_{fi})(x_{mj} - x_{fj})\right] \\ 
    &= \mathbb{E}\left[x_{mi}x_{mj} - x_{mi}x_{fj} - x_{fi}x_{mj} + x_{fi}x_{fj}\right] \\
    &= \mathbb{E}\left[x_{mi}x_{mj}\right] - \mathbb{E}\left[x_{mi}x_{fj}\right] - \mathbb{E}\left[x_{fi}x_{mj}\right] \nonumber  \\ 
    & + \mathbb{E}\left[x_{fi}x_{fj}\right]
\end{align*}

Note that the above term incorporates four terms of the kind $\mathbb{E}\left[x_{r_1}x_{r_2}\right]$ (where $r_1$ and $r_2$ are the two raters). Under our null hypothesis we need not distinguish between the raters based on their gender.  However, we must consider the event that both raters are the same. 

In the case the raters are not the same (and representing the two raters in a pair as $r_1$ and $r_2$), we have
\begin{align*}
    \mathbb{E}\left[x_{r_1}x_{r_2}\right] &= \mathbb{E}\left[ \mathbb{E}\left[x_{r_1} x_{r_2}  | r_1, r_2\right] \right]\\
    & = E\left[ \mathbb{E}\left[x_{r_1} | r_2 \right] \mathbb{E}\left[x_{r_1} | r_2\right] \right] \\
    & = \mathbb{E}\left[\mu_{r_1} \mu_{r_2} \right]
\end{align*}
The {\em rater mean} $\mu_r$ for any rater $r$ can be written in terms of the {\em global} mean rating $\mu_g$ and the deviation from the global mean as $\mu_r = \mu_g  + \epsilon_r$,  where $\epsilon_r$ is a zero-mean random variable. Thus, we can write
\begin{align}
  \mathbb{E}\left[x_{r_1}x_{r_2}\right]  & = \mathbb{E}\left[(\mu_{g} + \epsilon_{r_1}) (\mu_{g} + \epsilon_{r_2}) \right] \nonumber \\
    & = \mathbb{E} \left[\mu_{g}^2 + (\epsilon_{r_1} + \epsilon_{r_2})\mu_{g} + \epsilon_{r_1}\epsilon_{r_2}  \right] \nonumber \\
    & = \mathbb{E} \left[\mu_{g}^2\right]  \nonumber \\ 
    &= \mu_{g}^2 \label{eq:stat2}
\end{align}

In the case that the raters are the same, we have $r_1 = r_2 = r$ and we can write
\begin{align*}
    \mathbb{E}\left[x_{r_1}x_{r_2}\right] &= \mathbb{E}\left[ \mathbb{E}\left[x_1x_2  | r\right] \right]
\end{align*}
where $x_1$ and $x_2$ are two independent ratings produced by rater $r$.
The individual rating $x_1$ and $x_2$ can be written as the sum of the mean rating of the rater $\mu_r$ plus zero-mean perturbations $\epsilon_1$ and $\epsilon_2$ respectively.  Hence, we can write
\begin{align*}
    \mathbb{E}\left[x_{r_1}x_{r_2}\right] & = \mathbb{E}\left[ \mathbb{E} \left[(\mu_{r} + \epsilon_1) (\mu_{r} + \epsilon_2)\right] | r\right] \\
    & = \mathbb{E} \left[\mathbb{E}\left[\mu_{r}^2\right] + \mathbb{E}\left[\epsilon_1 + \epsilon_2\right]\mu_{r} + \mathbb{E}\left[\epsilon_1\epsilon_2  \right]\right] \\
    &= \mathbb{E}\left[\mu_{r} ^ 2\right]
\end{align*}
$\mu_r$, the mean rating for a rater, is itself a random variable. Let $\mu_r \sim P(\mu_g, \sigma_b^2)$, where, as before $\mu_g$ is the global mean and $\sigma_b^2$ is the {\em between-rater} variance. It follows that
\begin{equation}
    \mathbb{E}\left[x_{r_1}x_{r_2}\right] = \sigma_b^2 + \mu_{g}^2
\label{eq:stat3}
\end{equation}

We can now explicitly write out $\mathbb{E}\left[d_id_j\right]$ (from Equation \ref{eq:stat1}) for each of the possible conditions of rater repetition,  under $H_0$, using the values from Equations \ref{eq:stat2} and \ref{eq:stat3}. Let $m_j$ be the $j^{th}$ male rater and $f_j$ be the $j^{th}$ female rater:

We first note that 
\begin{align}
\mathbb{E}\left[d_id_j\right] &= \mathbb{E}\left[x_{mi}x_{mj}\right] - \mathbb{E}\left[x_{mi}x_{fj}\right] - \mathbb{E}\left[x_{fi}x_{mj}\right] \nonumber  \\ 
& + \mathbb{E}\left[x_{fi}x_{fj}\right]\label{eq:stat4}
\end{align}

In the case $m_i \neq m_j, f_i \neq f_j$ all raters are independent, and all four terms in the RHS of Equation \eqref{eq:stat4} are equal. Hence
\begin{align*}
\mathbb{E}\left[d_id_j\right] = 0
\end{align*}

In the case $m_i = m_j, f_i \neq f_j$, $\mathbb{E}\left[x_{mi}x_{fj}\right], \mathbb{E}\left[x_{fi}x_{mj}\right]$ and $\mathbb{E}\left[x_{fi}x_{fj}\right]$ are all equal and only $\mathbb{E}\left[x_{mi}x_{mj}\right]$ needs to be treated separately. Two of the three terms in the RHS of Equation \ref{eq:stat4} cancel out, leaving us with
\begin{align*}
\mathbb{E}\left[d_id_j\right] &= \mathbb{E}\left[x_{mi}x_{mj}\right] - \mathbb{E}\left[x_{mi}x_{fj}\right] \\
&=\sigma_b^2 + \mu_{g}^2 - \mu_{g}^2 \\
&=\sigma_b^2
\end{align*}
In the case $m_i \neq m_j, f_i = f_j$, the situation is the same as above.

In the case $m_i = m_j, f_i = f_j$
\begin{align*}
\mathbb{E}\left[d_id_j\right] &= \mathbb{E}\left[x_{mi}x_{mj}\right] - \mathbb{E}\left[x_{mi}x_{fj}\right] - \mathbb{E}\left[x_{fi}x_{mj}\right] \nonumber  \\ 
& + \mathbb{E}\left[x_{fi}x_{fj}\right] \\
&= \sigma_b^2 + \mu_{g}^2 - \mu_{g} ^ 2 - \mu_{g} ^ 2 + \sigma_b^2 + \mu_{g}^2\\
&= 2 \sigma_b^2 
\end{align*}


Given the above equations, we now summarize the four cases
\begin{align*}
\mathbb{E}\left[d_id_j\right]=
\begin{cases}
0,& m_i \neq m_j, f_i \neq f_j\\
\sigma_b^2, &  m_i = m_j, f_i \neq f_j\\
\sigma_b^2, &  m_i \neq m_j, f_i = f_j\\
 2 \sigma_b^2 , &  m_i = m_j, f_i = f_j
\end{cases}
\end{align*}

We assume that all the raters have the same mean rating.
We present the final variance term for the test statistic:

\begin{align*}
 \sigma_{\bar{x}}^2 = \frac{\sigma_{d}^2}{N} + M\frac{\sigma_{b}^2}{N^2}
\end{align*}
where $\sigma_{d}^2$ is the variance of the differences, $\sigma_{b}^2$ is the 
total between rater variance of the raters, $M$ is the total number of times any rater is repeated in the annotations. 

We estimate the between-rater variance as 
\begin{align*}
    \sigma_{b}^2 = \frac{1}{N_r-1}\sum_r\left(\mu_r - \mu_g\right)^2
\end{align*}
where $\mu_r$ is the mean rating of rater $r$, $\mu_g$ is the global mean rating, $N_r$ is the number of raters. Note that the above is a conservative estimate where we attempt to eliminate rater-specific biases in the variance by assigning equal weight to all raters. On our data it results in a larger estimate for $\sigma_b^2$ than what we obtain if we explicitly factor in the number of ratings by each rater.

Finally, the test statistic can be written as:
\begin{align*}
\text{tstat} & =  \frac{\bar{x} - \mu_x}{\sqrt{\sigma_{\bar{x}}^2}} \\ 
    & =  \frac{\bar{x} - \mu_x}{\sqrt{\frac{\sigma_{d}^2}{N} + M\frac{\sigma_{b}^2}{N^2}}}
\end{align*}

\subsection{Unpaired test}
We also test the hypothesis that the gender of the {\em speaker} affects the ratings assigned by raters of different genders. Here we will segregate all ratings assigned by (for instance) male raters by the gender of the speaker, and compare the mean rating of utterances by male speakers to that by female speakers.  Once again, we are faced with a mixed situation where the same rater may have rater multiple utterances by both male and female speakers, and the computation of the statistic must account for this variation. We will treat this as an unpaired test, but our derivation of the statistic naturally also accounts for all pairings that may result from having the same rater for male and female speakers.

The problem can be formulated as follows: given all the ratings from raters of a specific gender, we must determine if they perceive the speakers from two genders differently or not. The null hypothesis is that raters of a given gender does not perceive differently based on the gender of the speaker.  
Formally stated, let $\bar{x} = \bar{x}_f  - \bar{x}_m $, where $\bar{x}_f$ and $\bar{x}_m$ are the mean ratings assigned to male and female speakers respectively and are given by $\bar{x}_f = \frac{1}{N_f}\sum_{i\in U_f} x_i$ and $\bar{x}_m = \frac{1}{N_m}\sum_{i \in U_m} x_i$ where $U_f$ and $U_m$ are the sets of female and male-spoken utterances and $N_f$ and 
$N_m$ are their sizes. $x_i$'s (as before) are the ratings. 
The null and alternative hypothesis can be written like so:
\begin{align*}
H_0 : \mathbb{E}\left[\bar{x}\right] = 0 \\
H_A : \mathbb{E}\left[\bar{x}\right] \neq 0
\end{align*}

Under the $H_0$, the variance of $\bar{x}$ is obtained as
\begin{align*}
    \sigma_{\bar{x}}^2 & = \mathbb{E}\left[\bar{x}_f  - \bar{x}_m\right] \\
    &= \left(\frac{1}{N_f} + \frac{1}{N_m}\right)\sigma_{x}^2 + \\
&    \sigma_{b}^2\left(\frac{N^f_{r_1=r_2}}{N_f^2} + \frac{N^m_{r_1 = r_2}}{N_m^2} - 2\frac{N_{rm = rf}}{N_fN_m}\right)
\end{align*}
where $\sigma_{x}^2$ is the total variance of the ratings. 
$N^f_{r_1=r_2}$ and $N^m_{r_1=r_2}$ are the total number of pairs of female and male utterances respectively where the rater is the same, and $N_{rm = rf}$ is the number of combinations of the male and female utterances where the rater is the same. We leave the actual derivation of $\sigma_{\bar{x}}^2$ out for brevity. 

Finally the test statistic can be written as
\begin{align*}
\text{tstat}  =   \frac{\bar{x}}{\sqrt{\sigma_{\bar{x}}^2}} 
\end{align*}
where for $\sigma_{\bar{x}}^2$ we will use an unbiased estimator.

In both tests, the degrees of freedom equals the number of unique utterance-rater combinations, minus adjustment terms. However, since these numbers are large, the underlying distribution is effectively Gaussian. We will, however, conservatively assume 30 degrees of freedom -- the noted boundary where the difference between t and Gaussian pdfs becomes negligible, in computing our significance.

\section{Results}
We test the following hypotheses relating the gender of the {\em rater} to the gender of the {\em speaker} for each of the three dimensions of emotion (V, A, and D). 
\begin{itemize}
    \item $H_0(S|R=m)$: The gender of the speaker has no influence over the ratings assigned by male raters, i.e. male raters rate male and female speakers similarly. This requires our unpaired test, comparing the average rating assigned by male raters to female speakers to that assigned by them to male speakers. $H_0(S|R=f)$ is similarly defined.
    \item $H_0(R|S=m)$:  Raters of both genders rate male speakers similarly.  This requires our paired test. $H_0(R|S=f)$ is similarly defined.
    \item $H_0(R|S=a)$:  Raters of both genders rate utterances similarly. This  test does not distinguish the gender of the speaker (denoted by $S=a$, where "$a$" represents {\em all} genders). This can be performed by a paired test.
\end{itemize}

We employ two-sided t-tests using the t statistics described in the Section 5.  Table~\ref{tab:results_paired} states the test statistics and the confidence level with which the null hypothesis can be rejected.  The first column specifies the test. The remaining columns give the t-statistic and corresponding confidence value with which the null hypothesis can be rejected. In all cases the actual $p$ value was significantly lower than the confidence value.  
\begin{table}[ht]
\begin{tabular}{|c|c|c|c|c| } 
\hline
Hyp & V tstat($\alpha$) & A tstat($\alpha$) &  D tstat($\alpha$)  \\
\hline
$H_0(R|S=a)$ & 3.773(0.001) & 7.73($<$.001) & 6.332($<$.001) \\ 
$H_0(R|S=f)$ & 1.877(0.1) & 2.597(0.01) & 2.873(0.01)\\ 
$H_0(R|S=m)$  &  2.309(0.1) & 5.802($<$.001) & 5.93($<$.001)\\ 
\hline
\end{tabular}
\vspace{1em}
\caption{This tables shows the t-statistics  for the valence, arousal and dominance values across the three paired tests.
The $\alpha$ represents the minimum significance level at which that result is significant to reject the $H_0$
}
\label{tab:results_paired}
\end{table}
$H_0(S|R=m)$ and $H_0(S|R=f)$ are not shown in the table -- neither of them could be rejected at even the 0.1 level. There is no statistically significant difference in the manner in which male (or female) raters rate the speakers of the two genders.
It is important to note that the mean difference value between females and male raters is always positive. This means that females rate valence, arousal and dominance higher than males on average. This trend is also seen in other experiments performed in visual and speech domains. For example, \cite{montagne2005sex} found that men are poorer than women at perceiving emotion through facial expression. We hypothesize that the results achieved in one modality, like facial expression, may also reflect in the other modality, like speech. This is a very interesting result worth exploring. 

\section{Discussion}
The main focus of this study was to analyze crowdsourced data, where we dealt with challenges like the sparsity of the data and the dependence of the ratings over the multiple utterances because of the repetitions of raters. We also focused on understanding the difference between the perceived emotion in females and males from this data. The results show that significant difference exists between the genders. 

Analyzing the results presented in Table~\ref{tab:results_paired}, we can make several interesting observations. Firstly, male and female raters perceive emotion differently along all three dimensions, highest being in dominance. 
Secondly, the difference between female and male raters is less significant for female speakers than male speakers. 

The difference between males and females is not significant in the perception of valence of speakers of either gender individually; however in general, when speaker gender is not considered there remains a significant difference. 

Interestingly, the gender of the speaker does not appear to influence the ratings assigned by either male or female speakers.

There is a lot of potential in this data that can be used for emotion perception. In the future, we plan to account for the multiple annotations and rater pairs. This will provide us more data to work with. We also plan to explore how the sparse data be used in a better way. We plan to explore the correlation between the three emotion dimensions and study perception based on some knowledge of more than one emotion dimension. 

\section{Conclusion}

This paper has two major contributions. 
Firstly, we present a framework for presenting data in the wild and for crowd sourcing the annotations for emotional labels for speech. We make the data available to be used in research in this area or in general. The data will be useful for the study of perception.  
Secondly, we hypothesize that males and females perceive emotions differently. To test this hypothesis in the collected data, we formulate test statistics based on the structure of the data. The tests show promising and significant results. It is shown that with very high confidence females and males perceive emotion differently. This difference exists in all the three emotional dimensions, though highest in the dominance dimension. We also hypothesize that the subjects of any gender perceive the emotions of female and male speakers differently. However it was found that a significant difference does not exist.
We note that there exist some limitations to the data and this study. Crowd sourced data are noisy, sparse, and are not controlled and hence need to be dealt with very carefully. 
We make several assumptions regarding the data including that the raters have ``good faith'' and also that they represent the general population. We do perform filtering on the data to be consistent with our assumptions. 
Finally, we note several avenues of future work including dealing with the sparsity of the data in a better way and studying the correlation between the different dimensions of the emotions. 

%
%
%
%
%
\balance{}

\section{Supporting information}

\paragraph{S1 Appendix.}
\label{S1_Appendix}
{\bf VAD emotional state model.} The VAD (or PAD) emotional state model was introduced by Mehrabian and Russel \cite{mehrabian1974approach} to describe perceptions of physical environments in a three-factor model i.e. valence (or pleasure), arousal (or activation), and dominance. Valence represents how enjoyable an individual perceives the environment. Arousal represents the intensity of the stimulus. Dominance captures whether an individual feels in control of the environment or not. The values for each of the three components typically range from -1 to 1. The VAD factor-based theory states that every discrete emotion (e.g. happy, sad, etc.) has a mapping to the VAD space. The reverse may not be true. Since the values in the VAD model fall on a continuum, this makes it much more expressive than discrete emotions. This is why, we employ the VAD model in our study to capture the gender differences in perception of emotion in speech on a finer granularity. We are not the first study to use VAD model in the context of speech-based emotion study. There have been multiple other studies in the past such as \cite{vad1,vad2,vad3,vad4,vad5,vad6,vad7,vad8} which have worked on similar fronts mostly in the context of speech emotion recognition tasks.

\paragraph{S2 Appendix.}
\label{S2_Appendix}
{\bf AffectButton.} The AffectButton is a simple concept developed by Joost Boroekens \cite{broekens2013affectbutton}, and allows humans to point to the emotion recognized rather than naming it explicitly such as ``happy", ``ecstatic", ``sad", ``very sad", etc. Each point in the two dimensional space that represents the AffectButton then maps to coordinates in a 3-D space representing valence, arousal and dominance. The affective value triplets (valence, arousal, dominance) are represented by a dynamically changing rendered facial expression. Each facial expression is a combination of of eyebrow, eye and mouth configuration of the face, inspired by the Facial Action Coding System (FACS) \cite{ekman1997face}. The face is gender neutral and rudimentary to avoid any biases in the annotation process. The experiments conducted in the original paper conclude that the AffectButton is reliable, valid, takes less amount of screen space, and is quicker in terms of the annotation time. A detailed description of the AffectButton can be found in \cite{broekens2013affectbutton}. 

\begin{figure}[ht]
  \centering
  \scalebox{0.30}{
    \includegraphics{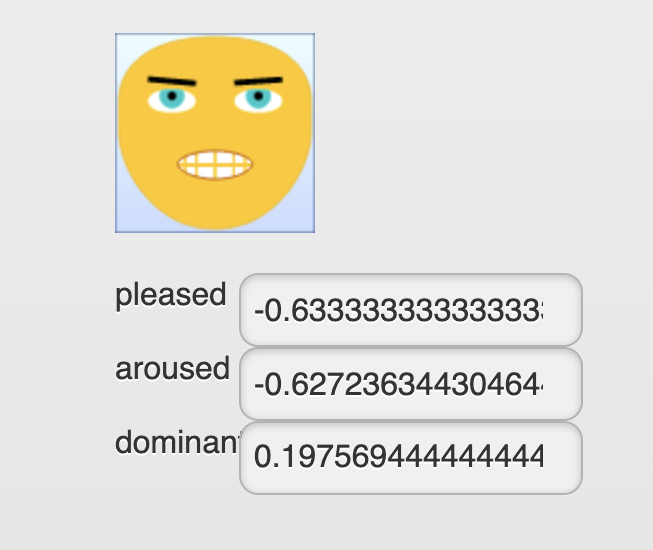}}
  \caption{Demonstration of the AffectButton's version 3.3 that we employ in our web application for the purposes of annotation collection.}
  \label{fig:affecticon}
\end{figure}

\section{Acknowledgment}

This material is based upon work funded and supported by the Department of Defense under Contract No. FA8702-15-D-0002 with Carnegie Mellon University for the operation of the Software Engineering Institute, a federally funded research and development center. DM19-0798.

\bibliographystyle{SIGCHI-Reference-Format}
\bibliography{sample}

\end{document}